\documentclass[10pt,twocolumn,letterpaper]{article}

\usepackage{cvpr}              
\usepackage[accsupp]{axessibility}
\usepackage{amsmath,graphicx,setspace}
\usepackage{amssymb}
\usepackage{algorithm}
\usepackage{algorithmic}
\usepackage{multirow}
\usepackage{bm}
\usepackage{subcaption}

\usepackage[dvipsnames]{xcolor}

\definecolor{cvprblue}{rgb}{0.21,0.49,0.74}
\usepackage[pagebackref,breaklinks,colorlinks,citecolor=cvprblue]{hyperref}

\def\paperID{3} 
\def\confName{CVPR}
\def\confYear{2024}

\title{Generative Dataset Distillation: Balancing Global Structure and Local Details}

\author{Longzhen Li, Guang Li, Ren Togo, Keisuke Maeda, Takahiro Ogawa, and Miki Haseyama\\
Hokkaido University\\
{\tt\small \{longzhen, guang, togo, maeda, ogawa, mhaseyama\}@lmd.ist.hokudai.ac.jp}
}

\begin{document}
\maketitle
\begin{abstract}
In this paper, we propose a new dataset distillation method that considers balancing global structure and local details when distilling the information from a large dataset into a generative model. Dataset distillation has been proposed to reduce the size of the required dataset when training models. The conventional dataset distillation methods face the problem of long redeployment time and poor cross-architecture performance. Moreover, previous methods focused too much on the high-level semantic attributes between the synthetic dataset and the original dataset while ignoring the local features such as texture and shape. Based on the above understanding, we propose a new method for distilling the original image dataset into a generative model. Our method involves using a conditional generative adversarial network to generate the distilled dataset. Subsequently, we ensure balancing global structure and local details in the distillation process, continuously optimizing the generator for more information-dense dataset generation.
\end{abstract}    
\section{Introduction}
The expansion of dataset sizes has notably propelled recent advancements in deep learning, especially within the field of computer vision~\cite{sengupta2020review}. However, the reliance on large datasets poses a challenge, as it often leads to considerable training expenses~\cite{alzubaidi2021review}. This issue can be addressed by two main approaches: data selection and dataset distillation.
Data selection involves selecting a subset of representative data from the original large dataset~\cite{zha2023data}. Although this approach can reduce the training cost, it risks losing critical information.
Dataset distillation, on the other hand, offers a more effective solution~\cite{wang2018datasetdistillation}. Rather than simply selecting existing data, it involves synthesizing a new and much smaller dataset that contains the important information of the original dataset. This approach can significantly reduce dataset size without substantially compromising performance. Moreover, dataset distillation offers a further advantage in terms of data privacy~\cite{li2023sharing, loo2024attack}.
\par
Dataset distillation, an emerging area of interest within the research community, has evolved significantly in its algorithms and applications~\cite{lei2023survey, yu2023review}. Initially, dataset distillation creates a smaller dataset to mimic the training performance of the original dataset using meta-learning~\cite{wang2018datasetdistillation}. Subsequent advancements introduced gradient matching methods, focusing on aligning the gradients of models trained on both the original and distilled datasets~\cite{zhao2021datasetcondensation}. It was further expanded with the introduction of distribution matching methods, which aim to adjust the smaller dataset’s distribution to closely resemble that of the original dataset~\cite{zhao2023distribution}. 
Recently, some dataset distillation methods based on matching the training trajectories have been proposed~\cite{cazenavette2022dataset, guo2024datm}. The training trajectory refers to the change in the model weights during the training process. The more similar the training trajectories of the teacher and student models are, the closer the performance of the student model will be to that of the teacher model.
\par
With the development of dataset distillation, the applications of dataset distillation have spaned various fields, including continual learning~\cite{yang2023efficient}, privacy preserving~\cite{li2020soft, li2022compressed}, and federated learning~\cite{song2023federated}. However, conventional dataset distillation methods often incur high redeployment costs because they rely on a fixed distillation ratio or the often-used image per class (IPC). Another challenge that the conventional dataset distillation method faces is the relatively poor cross-architecture performance. Distilled results on the small architecture will be hard to apply to a more complex architecture, which will lead to poor model generalization performance.
\par
To solve the above issues, a new dataset distillation method is introduced, namely distilling the dataset into a generative model (DiM)~\cite{wang2023dim}. Different from conventional methods, DiM distills the information of the whole dataset into a conditional generative adversarial network (GAN) model rather than images. This shift to model-based storage significantly enhances DiM's redeployment efficiency, as it eliminates the need for retraining when IPC or distillation ratios change, thus overcoming the limitation of conventional distillation methods. In the distillation process, DiM uses logit matching as the alignment strategy. Logit matching focuses on the image category, emphasizing global information and high-level semantic attributes. Therefore, logit matching aligns the distilled image with the original image in terms of their category, rather than exact visual details~\cite{tian2021relationship}. However, it overlooks finer details such as shapes and textures, which limits the distillation accuracy and performance on cross-architecture generalization.
\par
To address the issue of losing finer details, we propose a novel method that considers both global structure and local details. The motivation of our method is the integration of high-level semantic attributes with attention to local features that can improve the distillation process and hence generate more robust distilled datasets. The local features are extracted from the intermediate layers of the neural network, ensuring a more detailed representation of the data. Our method combines attention to both the broad, global aspects and the detailed, local features of images. Specifically, the proposed method introduces a novel loss function that simultaneously accounts for the final layer logits' discrepancies and the variance in local features contained in the intermediate network layers, ensuring a better distillation process. Therefore, our method offers a more comprehensive framework for dataset distillation, leading to more effective and accurate model training and better robustness. 
The effectiveness of our method has been verified through experiments on three benchmark datasets. Notably, our method's consideration of both global and local image aspects results in datasets that exhibit enhanced cross-architecture generalization capabilities, proving effective across various neural network types.

The contributions of this paper can be summarized as follows.
\begin{itemize}
    \item We propose a new dataset distillation method that considers both global structure and local details, which can generate more robust distilled datasets.
    \item 
    By distilling the information into a generative model instead of images, the proposed method significantly improved the redeployment efficiency, which prevented the high cost of re-optimization. 
    \item We verified the effectiveness of the proposed method on three benchmark datasets, including MNIST, Fashion MNIST, and CIFAR-10. The proposed method is also verified as having a better performance in cross-architecture generalization.
\end{itemize}

\section{Related Works}
In this section, we present an overview of various dataset distillation methods. These methods are categorized into three classes: performance matching, gradient matching, and distribution matching. The choice of method depends on factors such as dataset size, deployment time, and computational cost.

\subsection{Dataset Distillation Using Performance Matching}
First, we introduce dataset distillation methods using performance matching. The goal is to optimize distilled datasets so neural networks trained on them mirror the loss profiles of networks trained on original datasets. This parity in performance ensures that models leverage the distilled datasets as effectively as the original ones. Within the methods, it is separated into subclasses like meta-learning methods, exemplified by gradient-based hyperparameter optimization, and kernel ridge regression methods.
The inception of dataset distillation is first introduced by Wang et al.~\cite{wang2018datasetdistillation}, and they employed meta-learning paradigms to optimize model weights as functions of distilled images. 
Enhancements to this method have since emerged, introducing variations with flexible labels~\cite{bohdal2020flexible}, soft-label approaches~\cite{sucholutsky2021soft}, and the incorporation of parametrization~\cite{kim2022dataset} to improve distillation performance.
Meta-learning-based approaches employ backpropagation to compute the gradient of the validation loss on synthetic datasets, a process that requires bi-level optimization and can be computationally demanding, especially as the number of inner loops grows, leading to increased GPU memory usage~\cite{lorraine2020optimizing}. Limited inner loops can result in suboptimal optimization and performance issues, and scaling such methods to larger models presents further challenges~\cite{vicol2022implicit}.
However, kernel ridge regression methods like kernel inducing point (KIP) offer an alternative by enabling convex optimization, which yields a closed-form solution that obviates the need for exhaustive inner loop training~\cite{nguyen2021kip}.
Recent advancements in this domain have introduced methods that significantly enhance distillation efficiency and performance. These include leveraging infinitely wide convolutional networks~\cite{nguyen2021kipimprovedresults} and employing neural feature regression~\cite{zhou2022dataset}, each contributing to the evolution of dataset distillation methods.
\subsection{Dataset Distillation Using Gradient Matching}
Next, we introduce dataset distillation methods using gradient matching. Zhao et al. first proposed a gradient matching-based method termed dataset condensation~\cite{zhao2021datasetcondensation}.
Different from performance matching, which tunes the efficacy of models using synthetic datasets, gradient matching refines the performance of networks trained on both the original and synthetic datasets by aligning their training gradients.
Recent advancements have augmented gradient matching with strategies like differentiable data augmentation~\cite{zhao2021differentiatble}, enhancing training adaptability and efficacy. 
Furthermore, contrastive signaling~\cite{lee2022dataset} has been integrated to improve feature discrimination, and long-range trajectory matching~\cite{cazenavette2022dataset} has been employed to align gradient trajectories over extended training cycles. 
Additionally, approaches such as parameter pruning and self-adaptive parameter matching~\cite{li2023ddpp, li2024iadd} have been investigated to optimize distillation complexity, potentially boosting the efficiency and outcome of the distillation process.
\begin{figure*}[t]
        \centering
        \includegraphics[width=16cm]{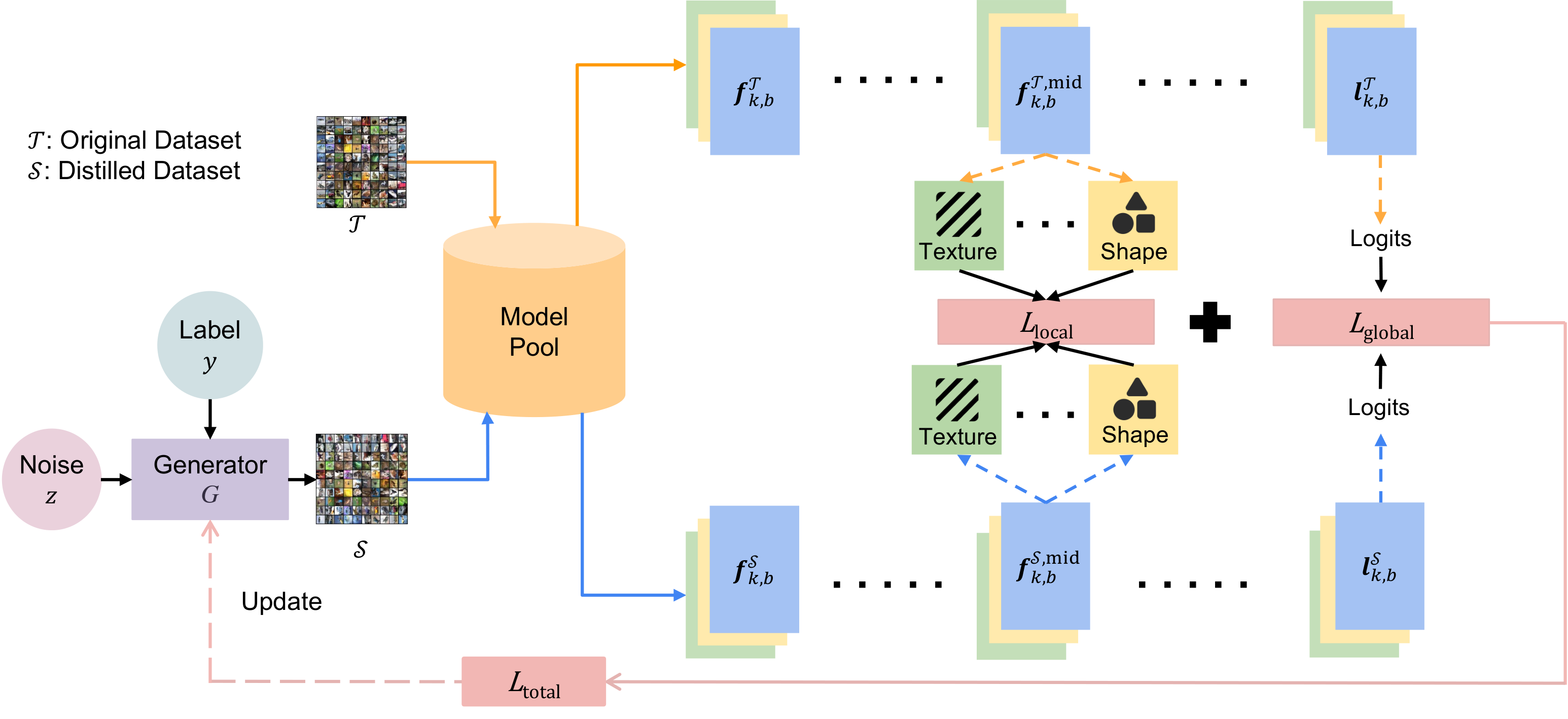}
        \caption{Overview of the distillation process. The goal is to train a generator that synthesizes images rich in information (referred to as distilled images), taking into account both global structure and local details.}
        \label{fig1}
\end{figure*}
\subsection{Dataset Distillation Using Distribution Matching}
Finally, we illustrate some dataset distillation methods using distribution matching. Distribution matching aims to synthesize data whose distribution closely aligns with that of the original data within a specified embedding space.
Zhao et al. first introduced the method of distribution matching, which utilizes the neural network's output embeddings, excluding the final linear layer~\cite{zhao2023distribution}. The objective is to minimize the distance between the mean vectors (centers) of synthetic and original data for each class.
Building on this, another method has been developed to align the attention across different layers of the network~\cite{sajedi2023datadam}. Although these distribution matching methods reduce synthesis costs and scale well with larger datasets, they require re-optimization with any change in the distillation ratio. This necessity for re-optimization can impact their efficiency in certain applications. 
For a deeper exploration of dataset distillation, the reader is directed to the latest survey papers~\cite{lei2023survey, yu2023review} or the Awesome Dataset Distillation project~\cite{li2022awesome}, which provide a thorough overview of the field.

\section{Methodology}
The proposed method aims to train a generator to synthesize information-condensed images. The proposed method includes three stages: conditional GAN training, dataset distillation via balancing global structure and local details, and the deployment stage.
\subsection{Conditional GAN Training}
Conventional GAN networks are used to generate visually realistic images. GAN networks generally include a generator $G$ and a discriminator $D$~\cite{goodfellow2014generative}. The working principle of GAN can be summarized as follows: 
\begin{equation}
\begin{split}
    L_{\textrm{GAN}} = \min_G \max_D V(D, G),
\end{split}
\end{equation}
\begin{equation}
\begin{split}
    \min_G \max_D V(D, G) = \,\, & \mathbb{E}_{\bm{x} \sim p(\bm{x})}[\log D(\bm{x})] \\ & \,\,+\,\,  \mathbb{E}_{\bm{z} \sim p(\bm{z})}[\log (1 - D(G(\bm{z})))],
\end{split}
\end{equation}
where $\bm{x}$ represents real data and $\bm{z}$ represents random noise. During the GAN training process, the generator $G$ and discriminator $D$ compete with each other. The generator $G$ attempts to generate more realistic images and the discriminator $D$ tries to distinguish between the generated and real data.
\par
Different variants of the GAN network were later developed, and in the proposed method, we chose conditional GAN as the generative model to generate the distilled data. Compared with conventional GAN networks, conditional GAN introduces specific information into the input to generate images~\cite{mirza2014conditional}. In the proposed method, we use labels as specific information. The training process of conditional GAN can be summarized as follows: 
\begin{equation}
    \begin{split}
        L_{\textrm{CGAN}} = \min_G \max_D V(D, G),
    \end{split}
\end{equation}
\begin{equation}
\begin{split}
    \min_G \max_D V(D, G) = \,\, & \mathbb{E}_{\bm{x} \sim p(\bm{x})}[\log D(\bm{x} | \bm{y})] \\ & \,\,+\,\, \mathbb{E}_{\bm{z} \sim p(\bm{z})}[\log (1 - D(G(\bm{z} | \bm{y})))],
\end{split}
\end{equation}
where $\bm{y}$ represents the introduced labels.
\par
The first step of the proposed method is to train a generator that can generate visually realistic images. The process for the trained generator in conditional GAN to generate discriminative images can be summarized as follows:
\begin{equation}
    \mathcal{S}=G \left ( \left [ \bm{z}\oplus \bm{y}  \right ];\mathcal{W}  \right ),
\end{equation}
where $\bm{z}$ is the random noise and $\bm{y}$ is the corresponding label. $\oplus$ denotes the concatenation operation. $\mathcal{S}$ represents the synthetic dataset and $\mathcal{W}$ is the parameter of generator $G$.
The synthetic dataset $\mathcal{S}$ is initially generated by the generator $G$ and subsequently optimized through the dataset distillation process, which distills the information from the original dataset $\mathcal{T}$. The optimizing method introduced in the next section will update $G$ to achieve better performance to generate more efficient images that contain more valuable information. 
\par
The most important difference between conventional dataset distillation methods and the proposed method is that the former ultimately saves the distilled images, which means distilling the information into images, whereas the goal of the proposed method is to save the trained generator, which means distilling the information into a generative model. The trained generator can be used to generate any number of distilled images, which saves a lot of redeployment costs.

\subsection{Dataset Distillation via Balancing Global Structure and Local Details}
The obtained conditional GAN generator can enable the synthesis of visually convincing images. However, the synthetic dataset $\mathcal{S}$ often lacks the compressed information in the original dataset $\mathcal{T}$, hindering the performance of downstream tasks. Our method tackles this limitation by focusing on the optimization of the generator, aiming to enhance its capability to produce distilled data that goes beyond simple visual authenticity, capturing a more potent and discerning representation of the important information.

As shown in Fig.~\ref{fig1}, we use a random initial model from the model pool to match the global structure and local details between the original dataset $\mathcal{T}$ and the synthetic dataset $\mathcal{S}$. Then it continuously optimizes the generator by minimizing the loss between the synthetic dataset and the original dataset, thereby generating data that is more efficient for downstream classification tasks. Unlike conventional dataset distillation methods, we use a model pool that contains multiple convolutional neural networks and randomly select one from them to perform the matching between the original dataset $\mathcal{T}$ and the synthetic dataset $\mathcal{S}$. With the model pool, we can improve the robustness and generalization performance of dataset distillation. By using different models to perform matching steps, the features of the original dataset $\mathcal{T}$ are more fully utilized. The use of a model pool makes the proposed method have better cross-architecture stability, avoiding overfitting to specific architectures and making our method more robust.

In our method, the matching of synthetic dataset and original dataset can be divided into two parts, the matching of global structure and the matching of local details. The matching of global structure aims to analyze whether the synthetic dataset is consistent with the original dataset in terms of high-level semantic information, such as categories. The global loss is obtained by comparing the high-level semantic information of the original dataset and the synthetic dataset. When comparing the global information, we use logical matching to compare whether the synthetic dataset has achieved similar logits to the original dataset. This step uses the output from the last layer of a randomly selected model. The global loss can be defined as follows:
\begin{equation}
L_{\textrm{global}} = \sum_{k=1}^{K}\sum_{b=1}^{B}({\mathbf{l}_{k,b}^{\mathcal{S}}}-{\mathbf{l}_{k,b}^{\mathcal{T}}})^2,
\end{equation}
where $B$ and $K$ denote the batch size and the number of categories, respectively. $\mathbf{l}^{\mathcal{S}}$ and $\mathbf{l}^{\mathcal{T}}$ represent the output logits of the synthetic dataset and original dataset, respectively.

However, focusing only on global information will cause the loss of valuable detailed information in the data. Therefore, we further perform matching on local features such as texture and shape. In this way, the synthetic dataset $\mathcal{S}$ contains more valuable detailed information.
When calculating the local loss, we propose using feature matching and selecting information from intermediate layers to compare the matching degree between the original data and the synthetic dataset in terms of texture, shape, and other detail aspects. The local loss can be calculated as follows:
\begin{equation}
L_{\textrm{local}} = \sum_{k=1}^{K}\sum_{b=1}^{B}({\mathbf{f}_{k,b}^{\mathcal{S},\textrm{mid}}}-{\mathbf{f}_{k,b}^{\mathcal{T},\textrm{mid}}})^2,
\end{equation}
where $\textrm{mid}$ denotes an intermediate layer of the randomly selected network. $\mathbf{f}^{\mathcal{S}}$ and $\mathbf{f}^{\mathcal{T}}$ represent the output features of the synthetic dataset and original dataset, respectively.

The total loss function of the proposed method is a combination of global loss $L_{\textrm{global}}$, local loss $L_{\textrm{local}}$, and conditional GAN loss $L_{\textrm{CGAN}}$. We also defined $\omega_g$ and $\omega_l$ to represent the weights of global loss and local loss. The calculation of total loss $L_{\textrm{total}}$ can be summarized as follows: 
\begin{equation}
L_{\textrm{total}} = \omega_gL_{\textrm{global}} + \omega_lL_{\textrm{local}} + L_{\textrm{CGAN}}.
\end{equation}
During the dataset distillation stage, the optimization process focuses on minimizing the total loss function. Through this minimization process, the generator progressively improves its ability to generate data similar to the desired target distribution. This minimization process can be summarized as follows:
\begin{equation}
\mathcal{W}^{\ast} = \underset{\mathcal{W}}{\arg\min}\,L_{\textrm{total}},
\end{equation}
where $\mathcal{W}^{\ast}$ is the optimized parameters of the generator $G$.
The proposed method ensures balancing global structure and local details of the original dataset and synthetic dataset during the dataset distillation process as much as possible by matching them and making the synthetic dataset contain as much detailed information as possible, thereby generating distilled datasets for downstream tasks.
\subsection{Deployment Stage}
After the above optimization, the generator cannot only generate visually realistic images but also generate distilled images. These distilled images contain more key information that is helpful for downstream tasks such as recognition and classification.
Therefore, during the deployment phase, we provide various random noises $\mathbf{z}$ and corresponding labels $\mathbf{y}$ to the generator and use the generator to dynamically generate various distilled dataset $\mathcal{S}^{\ast}$ as follows: 
\begin{equation}
\mathcal{S}^{\ast} = G \left ( \left [ \mathbf{z}\oplus \mathbf{y}  \right ];\mathcal{W}^{\ast}  \right ).
\end{equation}

This distilled dataset can be used to serve as the alternative for the original dataset to effectively reduce the volume of the dataset.
Moreover, since we saved the trained generator, the information of the whole dataset was distilled into the generative model during this process, rather than static images. Therefore, when we apply the new proposed method to other architectures or change the distillation ratio, there is no need to retrain the model. This improves the efficiency of redeployment a lot. Algorithm~\ref{alg1} summarizes the proposed method. The generative dataset distillation method trains a generator $G$ of conditional GAN first. Then the global-local coherence of the original and synthetic was matched. Finally, the generator $G$ is updated to generate a more efficient distilled dataset.

\begin{table*}[t]
    \centering
    \small
    \caption{Comparation with data selection methods and SOTA dataset distillation methods on three benchmark datasets.}
    \label{tab1}
    \begin{tabular}{cc|ccc|ccc|ccc}
    \hline
    \multicolumn{2}{c|}{Dataset} &\multicolumn{3}{c|}{MNIST} &\multicolumn{3}{c|}{Fashion MNIST} &\multicolumn{3}{c}{CIFAR-10} \\
    \multicolumn{2}{c|}{IPC} & 1 & 10 & 50 & 1 & 10 & 50 & 1 & 10 & 50\\\hline\hline
    \multicolumn{2}{c|}{Random~\cite{zhao2021datasetcondensation}} 
    & 64.9$\pm$3.5 & 95.1$\pm$0.9 & 97.9$\pm$0.2 & 51.4$\pm$3.8 & 73.8$\pm$0.7 & 82.5$\pm$0.7 & 14.4$\pm$2.0 & 26.0$\pm$1.2 & 43.4$\pm$1.0 \\
    \multicolumn{2}{c|}{Herding~\cite{chen2010super}} 
    & 89.2$\pm$1.6 & 93.7$\pm$0.3 & 94.8$\pm$0.2 & 67.0$\pm$1.9 & 71.1$\pm$0.7 & 71.9$\pm$0.8 & 21.5$\pm$1.3 & 31.6$\pm$0.7 & 40.4$\pm$0.6 \\
    \multicolumn{2}{c|}{K-Center~\cite{chierichetti2017fair}} 
    & 89.3$\pm$1.5 & 84.4$\pm$1.7 & 97.4$\pm$0.3 & 66.9$\pm$1.8 & 54.7$\pm$1.5 & 68.3$\pm$0.8 & 21.5$\pm$1.3 & 14.7$\pm$0.9 & 27.0$\pm$1.4 \\
    \multicolumn{2}{c|}{Forgetting~\cite{toneva2019empirical}} 
    & 35.5$\pm$5.6 & 68.1$\pm$3.3 & 88.2$\pm$1.2 & 42.0$\pm$5.5 & 53.9$\pm$2.0 & 55.0$\pm$1.1 & 13.5$\pm$1.2 & 23.3$\pm$1.0 & 23.3$\pm$1.1 \\\hline\hline
    \multicolumn{2}{c|}{DC~\cite{zhao2021datasetcondensation}} 
    & 91.7$\pm$0.5 & 97.4$\pm$0.2 & 98.8$\pm$0.2 & 70.5$\pm$0.6 & 82.3$\pm$0.4 & 83.6$\pm$0.4 & 28.3$\pm$0.5 & 44.9$\pm$0.5 & 53.9$\pm$0.5 \\
    \multicolumn{2}{c|}{DSA~\cite{zhao2021differentiatble}}
    & 88.7$\pm$0.6 & 97.8$\pm$0.1 & 99.2$\pm$0.1 & 70.6$\pm$0.6 & 84.6$\pm$0.3 & 88.7$\pm$0.2 & 28.8$\pm$0.7 & 52.1$\pm$0.5 & 60.6$\pm$0.5 \\
    \multicolumn{2}{c|}{DM~\cite{zhao2023distribution}}
    & 89.9$\pm$0.8 & 97.6$\pm$0.1 & 98.6$\pm$0.1 & 71.5$\pm$0.5 & 83.6$\pm$0.2 & 88.2$\pm$0.1 & 26.5$\pm$0.4 & 48.9$\pm$0.6 & 63.0$\pm$0.4 \\
    \multicolumn{2}{c|}{EGM~\cite{jiang2022delving}}
    & 91.9$\pm$0.4 & 97.9$\pm$0.2 & 98.6$\pm$0.1 & 71.4$\pm$0.4 & 85.4$\pm$0.3 & 87.9$\pm$0.2 & 30.0$\pm$0.6 & 50.2$\pm$0.6 & 60.0$\pm$0.4 \\
    \multicolumn{2}{c|}{CAFE~\cite{wang2022cafe}}
    & 93.1$\pm$0.3 & 97.5$\pm$0.1 & 98.9$\pm$0.2 & 73.7$\pm$0.7 & 83.0$\pm$0.4 & 88.2$\pm$0.3 & 31.6$\pm$0.8 & 50.9$\pm$0.5 & 62.3$\pm$0.4 \\
    \multicolumn{2}{c|}{KIP~\cite{nguyen2021kipimprovedresults}}
    & 90.1$\pm$0.1 & 97.5$\pm$0.0 & 98.3$\pm$0.1 & 70.6$\pm$0.6 & 84.6$\pm$0.3 & 88.7$\pm$0.2 & 49.9$\pm$0.2 & 62.7$\pm$0.3 & 68.6$\pm$0.2 \\
    \multicolumn{2}{c|}{MTT~\cite{cazenavette2022dataset}}
    & 91.4$\pm$0.9 & 97.3$\pm$0.1 & 98.5$\pm$0.1 & 75.1$\pm$0.9 & 87.2$\pm$0.3 & 88.3$\pm$0.1 & 46.3$\pm$0.8 & 65.3$\pm$0.7 & 71.6$\pm$0.2 \\
    \multicolumn{2}{c|}{FRePo~\cite{zhou2022dataset}}
    & 93.8$\pm$0.6 & 98.4$\pm$0.1 & 99.2$\pm$0.1 & 75.6$\pm$0.5 & 86.2$\pm$0.3 & 89.6$\pm$0.1 & 46.8$\pm$0.7 & 65.5$\pm$0.6 & 71.7$\pm$0.2 \\\hline\hline
    \multicolumn{2}{c|}{CGAN~\cite{mirza2014conditional}}
    & 96.1$\pm$0.7 & 97.8$\pm$0.3 & 98.4$\pm$0.3 & 81.5$\pm$0.5 & 84.0$\pm$0.2 & 86.3$\pm$0.3 & 46.4$\pm$1.2 & 62.7$\pm$0.9 & 68.1$\pm$0.8 \\
    \multicolumn{2}{c|}{DiM~\cite{wang2023dim}}
    & 96.5$\pm$0.6 & 98.6$\pm$0.2 & \bfseries{99.2$\pm$0.2} & 84.5$\pm$0.4 & 88.2$\pm$0.2 & 89.8$\pm$0.1 & 51.3$\pm$1.0 & 66.2$\pm$0.5 & 72.6$\pm$0.4 \\
    \multicolumn{2}{c|}{Ours}
    & \bfseries{97.3$\pm$0.3} & \bfseries{98.8$\pm$0.2} & 99.1$\pm$0.1 & \bfseries{85.5$\pm$0.2} & \bfseries{88.6$\pm$0.1} & \bfseries{90.1$\pm$0.1} & \bfseries{52.3$\pm$0.6} & \bfseries{66.7$\pm$0.3} & \bfseries{73.1$\pm$0.2} \\\hline\hline
    \multicolumn{2}{c|}{Original Dataset}
    & & 99.6$\pm$0.0 & & & 93.5$\pm$0.1 & & & 84.8$\pm$0.1 & \\\hline
    \end{tabular}
\end{table*}
\begin{algorithm}[t]
    \caption{Generative dataset distillation considering global-local coherence}
    \label{alg1}
    \begin{algorithmic}[1]
    \REQUIRE 
    $G$: generator of conditional GAN;
    $\mathcal{W}$: parameter of generator $G$;
    $z$: random noises;
    $y$: random label;
    $\varepsilon$: learning rate
    \ENSURE
    $\mathcal{W}^{\ast}$: the optimized parameters; 
    $\mathcal{S}^{\ast}$: the distilled dataset
    \\
    \STATE
    Train a conditional GAN
    \FOR{ each epoch $c$ = 1 to C }
    \FOR{ each interaction $i$ = 1 to I }
    \STATE
    Calculate the conditional GAN loss $L_{\textrm{CGAN}}$
    \STATE
    Update the parameter of $G$:
    \STATE
    $\mathcal{W}=\mathcal{W}-\varepsilon \frac{\partial L_{\textrm{CGAN}} }{\partial \mathcal{W}}  $
    \ENDFOR
    \ENDFOR
    \STATE
    Matching the global-local coherence of the original dataset and synthetic dataset:
    \FOR{ each epoch $e$ = 1 to E }
    \FOR{ each interaction $i$ = 1 to I }
    \STATE
    Calculate the total loss:
    \STATE
    $L_{\textrm{total}} = \omega_gL_{\textrm{global}} + \omega_lL_{\textrm{local}} + L_{\textrm{CGAN}}$
    \STATE
    Update the parameter of $G$:
    \STATE
    $\mathcal{W}=\mathcal{W}-\varepsilon \frac{\partial L_{\textrm{total}} }{\partial \mathcal{W}}  $
    \ENDFOR
    \ENDFOR
    \STATE
    Obtain the optimized parameter of $G$:
    \STATE
    $\mathcal{W}^{\ast} = \underset{\mathcal{W}}{\arg\min}\, L_{\textrm{total}}$
    \STATE
    Generate the distilled dataset $\mathcal{S}^{\ast}$:
    \STATE
    $\mathcal{S}^{\ast} = G \left ( \left [ \bm{z}\oplus \bm{y}  \right ];\mathcal{W}^{\ast}  \right )$
    \end{algorithmic}
\end{algorithm}
\section{Experiments}
\subsection{Experimental Settings}
\begin{figure*}[t]
    \centering
    \subcaptionbox{MNIST}{
        \includegraphics[width=5.5cm]{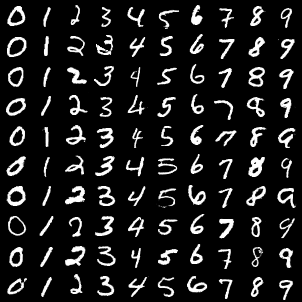}
    }
    \subcaptionbox{Fashion MNIST}{
        \includegraphics[width=5.5cm]{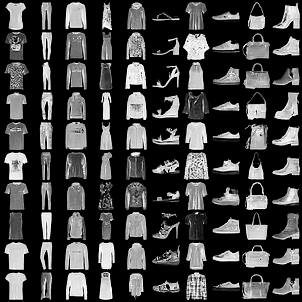}
    }
    \subcaptionbox{CIFAR-10}{
        \includegraphics[width=5.5cm]{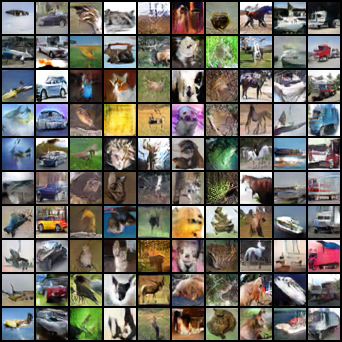}
    }
    \caption{Distilled MNIST, Fashion MNIST, and CIFAR-10 datasets with IPC = 10.}
    \label{fig2}
\end{figure*}
We used three benchmark datasets (MNIST, Fashion MNIST, and CIFAR-10) in the experiments for comparison with other methods.
They all have 10 classes and the resolution of images in the three datasets is all 32 $\times$ 32.
For comparative methods, we used seven SOTA dataset distillation methods, including dataset condensation (DC)~\cite{zhao2021datasetcondensation}, differentiable siamese augmentation (DSA)~\cite{zhao2021differentiatble}, distribution matching (DM)~\cite{zhao2023distribution}, aligning features (CAFE)~\cite{wang2022cafe}, kernel inducing point (KIP)~\cite{nguyen2021kipimprovedresults}, matching training trajectories (MTT)~\cite{cazenavette2022dataset}, and neural feature regression with pooling (FrePo)~\cite{zhou2022dataset}.
We also compared with baseline method CGAN~\cite{mirza2014conditional} and the generative-based dataset distillation method DiM~\cite{wang2023dim}.

To improve the generalization performance and avoid over-reliance on a single network, when optimizing the generator, we applied the model pool to get the randomly initialized model. The model pool has several models such as ConvNet3~\cite{gidaris2018dynamic}, ResNet10, and ResNet18~\cite{he2016deep}. Randomly selected models are used to match the global and local features of the synthetic dataset and the original dataset. When matching local features between two datasets, we focus on specific intermediate layers that demonstrate a superior ability to extract local features, such as the second layer within ResNet~\cite{yuan2022pre}.
We conducted three experiments to verify the effectiveness of the proposed method, including benchmark comparison, cross-architecture generalization, and hyperparameter ablation study.
All the experimental results are average accuracy and standard deviation of five networks trained from scratch on the distilled dataset and tested on the original dataset.
\subsection{Benchmark Comparison}
In this subsection, we verify the effectiveness of the proposed method by comparing it with other methods on three benchmark datasets, i.e., MNIST, Fashion MINIST, and CIFAR-10.
We designed three sets of experiments for each dataset. Each set applies IPC = 1, 10, and 50, respectively. ConvNet3 was used as the test model.
From Table~\ref{tab1}, we can see that our method achieves better performance under majority settings and shows better stability. In most experiments, the accuracy has been improved by about 0.5\%, especially when IPC = 1, the proposed method improved the accuracy by about 1\% and improved the stability. Figure~\ref{fig2} shows the visualization results on distilled MNIST, fashionMNIST, and CIFAR-10 datasets obtained using the proposed method. Based on the main experimental and visualization results, we can see that the proposed method can improve the performance of generative dataset distillation while maintaining the visual authenticity of the distilled dataset. 
\subsection{Cross-architecture Generalization}
In this subsection, we verify the effectiveness of our method in cross-architecture generalization. Cross-architecture means using distilled images generated by some architectures and testing on other architectures.
In our experiments, ConvNet3 and ResNet18 were selected as matching models when optimizing the generator $G$. To validate the generalization performance on other architectures, we used AlexNet~\cite{krizhevsky2012imagenet} and VGG11~\cite{simonyan2015very}. These architectures were trained on the distilled dataset and tested on the original dataset. We set the IPC = 10 to keep the same setting as the previous methods to make a fair comparison. 
Table~\ref{tab2} shows that the proposed method outperforms conventional dataset distillation methods in terms of cross-architecture generalization. The distilled dataset demonstrates higher accuracy across different architectures. In comparison with the DiM method, the distillation data derived from the proposed method shows enhancements in performance on various architectures and exhibits better stability.
\begin{table}[t]
    \centering
    \footnotesize
    \caption{Cross-architecture generalization ability comparation on CIFAR-10 dataset with IPC = 10.}
    \label{tab2}
    \begin{tabular}{lcccc}
    \hline
    Method & ConvNet3 & ResNet18 & AlexNet & VGG11 \\\hline\hline
    DSA~\cite{zhao2021differentiatble} & 52.1$\pm$0.4 & 42.8$\pm$1.0 & 35.9$\pm$1.3 & 43.2$\pm$0.5 \\
    KIP~\cite{nguyen2021kipimprovedresults} & 47.6$\pm$0.9 & 36.8$\pm$1.0 & 24.4$\pm$3.9 & 42.1$\pm$0.4 \\
    MTT~\cite{cazenavette2022dataset} & 64.3$\pm$0.7 & 46.4$\pm$0.6 & 34.2$\pm$2.6 & 50.3$\pm$0.8 \\
    FRePo~\cite{zhou2022dataset} & 65.5$\pm$0.4 & 57.7$\pm$0.7 & 61.9$\pm$0.7 & 59.4$\pm$0.7 \\
    DiM~\cite{wang2023dim} & 66.2$\pm$0.5 & 69.2$\pm$0.3 & 67.3$\pm$0.9 & 66.8$\pm$0.5 \\
    Ours & \bfseries{66.7$\pm$0.3} & \bfseries{71.6$\pm$0.2} & \bfseries{68.0$\pm$0.3} & \bfseries{67.4$\pm$0.6} \\
    \hline
    \end{tabular}
\end{table}
\begin{figure}[t]
        \centering
        \includegraphics[width=6cm]{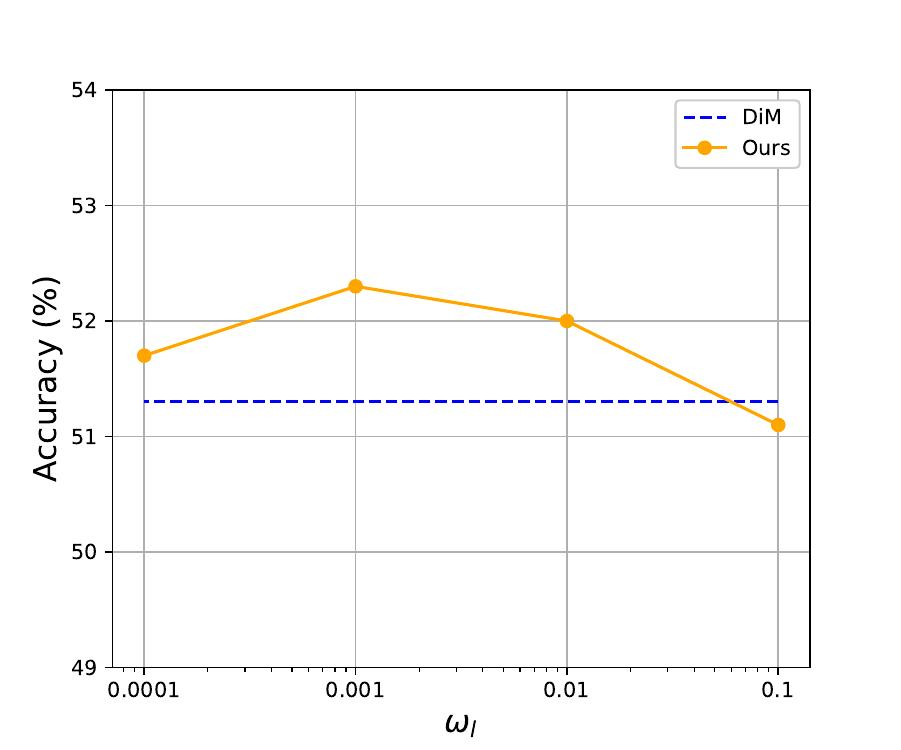}
        \caption{Ablation study of $\omega_l$ on CIFAR-10 dataset with IPC = 1.}
        \label{fig3}
\end{figure}
\subsection{Hyperparameter Ablation Study}
Since DiM has proved that the weight of global loss $\omega_g$ = 0.01 leads to the best performance. Hence, we used the same value of $\omega_g$ and set up an experiment on CIFAR-10 with IPC = 1 to explore the impact of the weight of local loss $\omega_l$. As shown in Fig.~\ref{fig3}, when the weight of the local loss $L_{\textrm{local}}$ was set to 0.001, the proposed method achieved the highest average accuracy. When the local loss weight is too large, it reduces the impact of global loss $L_{\textrm{global}}$ and conditional GAN loss $L_{\textrm{CGAN}}$, thereby reducing the accuracy, while a local loss weight that is too small will not allow the generator to effectively learn the local features. Although DiM has shown the best value of $\omega_g$, the impact of the different $\omega_g$ values is still worth exploring in future works. 
\section{Conclusion}
This paper has proposed a novel dataset distillation method. During the dataset distillation process, the proposed method takes into account both the global structure and local details, thereby ensuring that high-level semantic information and mid-level feature information are simultaneously distilled into the generative model. Experimental results show that the proposed method outperforms other SOTA dataset distillation methods on three benchmark datasets.

\subsection*{Acknowledgement}
This research was supported in part by JSPS KAKENHI Grant Numbers JP21H03456, JP23K11211, and JP23K11141.

{
    \small
    \bibliographystyle{ieeenat_fullname}
    \bibliography{main}
}
\end{document}